\documentclass[letterpaper, 10 pt, conference]{ieeeconf}  

\IEEEoverridecommandlockouts                              

\usepackage{graphics} 
\usepackage{graphicx}
\usepackage{multirow}
\usepackage{amsmath,xparse}
\usepackage{makecell}
\usepackage[dvipsnames]{xcolor}
\usepackage{cite}
\usepackage{diagbox}
\usepackage{dirtytalk}

\newcommand{\etal}{\textit{et al.}}

\title{\LARGE \bf
PillarFlow: End-to-end Birds-eye-view Flow Estimation \\ for Autonomous Driving
}

\author{
Kuan-Hui Lee, Matthew Kliemann, Adrien Gaidon, Jie Li, Chao Fang, Sudeep Pillai, Wolfram Burgard 
\thanks{All the authors are with Toyota Research Institute, USA {\tt\footnotesize \{first name\}.\{last name\}@tri.global}}
}

\begin{document}

\maketitle
\thispagestyle{empty}
\pagestyle{empty}

\begin{abstract}

In autonomous driving, accurately estimating the state of surrounding obstacles is critical for safe and robust path planning.
However, this perception task is difficult, particularly for generic obstacles/objects, due to appearance and occlusion changes.
To tackle this problem, we propose an end-to-end deep learning framework for LIDAR-based flow estimation in bird's eye view (BeV).
Our method takes consecutive point cloud pairs as input and produces a 2-D BeV flow grid describing the dynamic state of each cell.
The experimental results show that the proposed method not only estimates 2-D BeV flow accurately but also improves tracking performance of both dynamic and static objects.
%
%

\end{abstract}

\section{INTRODUCTION}
Robust and accurate perception of the surrounding environment is critical for downstream modules of an autonomous driving system, such as prediction and planning. In practice, perceptual errors result in braking and swerving maneuvers that are unsafe and uncomfortable.
Most autonomous driving systems utilize a \say{detect-then-track} approach to perceive the state of objects in the environment \cite{badue2019self}.
This approach has strongly benefited from recent advancements in 3-D object detection \cite{chen2017multi,ku2018joint,yang2018pixor,zhou2018voxelnet,lang2019pointpillars} and state estimation \cite{held2014combining,weng2019baseline,chiu2020probabilistic}.
However, making this architecture robust is an open challenge, as it relies on the geometric consistency of object detections over time. In particular, detect-then-track must handle (at least) the following errors:
\begin{itemize}
\item \textit{false negatives}, i.e. missed detections;
\item \textit{false positives}, i.e.~hallucinated objects~\cite{buehler2020driving};
\item \textit{out of ontology objects}, i.e.~object categories not labeled at training time and hence not detected or recognized by the detector, e.g., road debris, wild animals;
\item \textit{incorrect motion estimation} due to large viewpoint changes, occlusions, and
temporally inconsistent detections in dynamic scenes;
\item \textit{incorrect associations} due to compounding errors in tracking or poor state initialization.
\end{itemize}

To tackle these challenges, we present a LIDAR-based scene motion estimator that is \emph{decoupled} from object detection and thus complementary.
Our method takes two consecutive full LIDAR point cloud sweeps as input. Each LIDAR sweep is encoded into a discretized 2-D BeV representation of learned feature vectors, a.k.a.~``pillars"~\cite{lang2019pointpillars}.
Then, we learn an optical flow network, adapted from Sun \etal~\cite{sun2018pwc}, to locally match pillars between the two consecutive BeV feature grids. The whole architecture is learned end-to-end and the final output is a 2-D flow vector for each grid cell.

Our approach relies on a 2-D BeV representation over a 3-D or projective representation (depth image) for multiple reasons.
First, for autonomous driving, we primarily care about motion occurring on the road and adjacent surfaces, especially for motion planning.
Second, this Euclidean representation allows us to design the network architecture to leverage spatial priors on relative scene motion.
Finally, a 2-D representation is more computational efficient compared to volumetric approaches and facilitates the sharing of representations with an object detector running in parallel to our object-agnostic flow network.




\begin{figure}[t]
\vspace{2mm}
\includegraphics[width=\columnwidth]{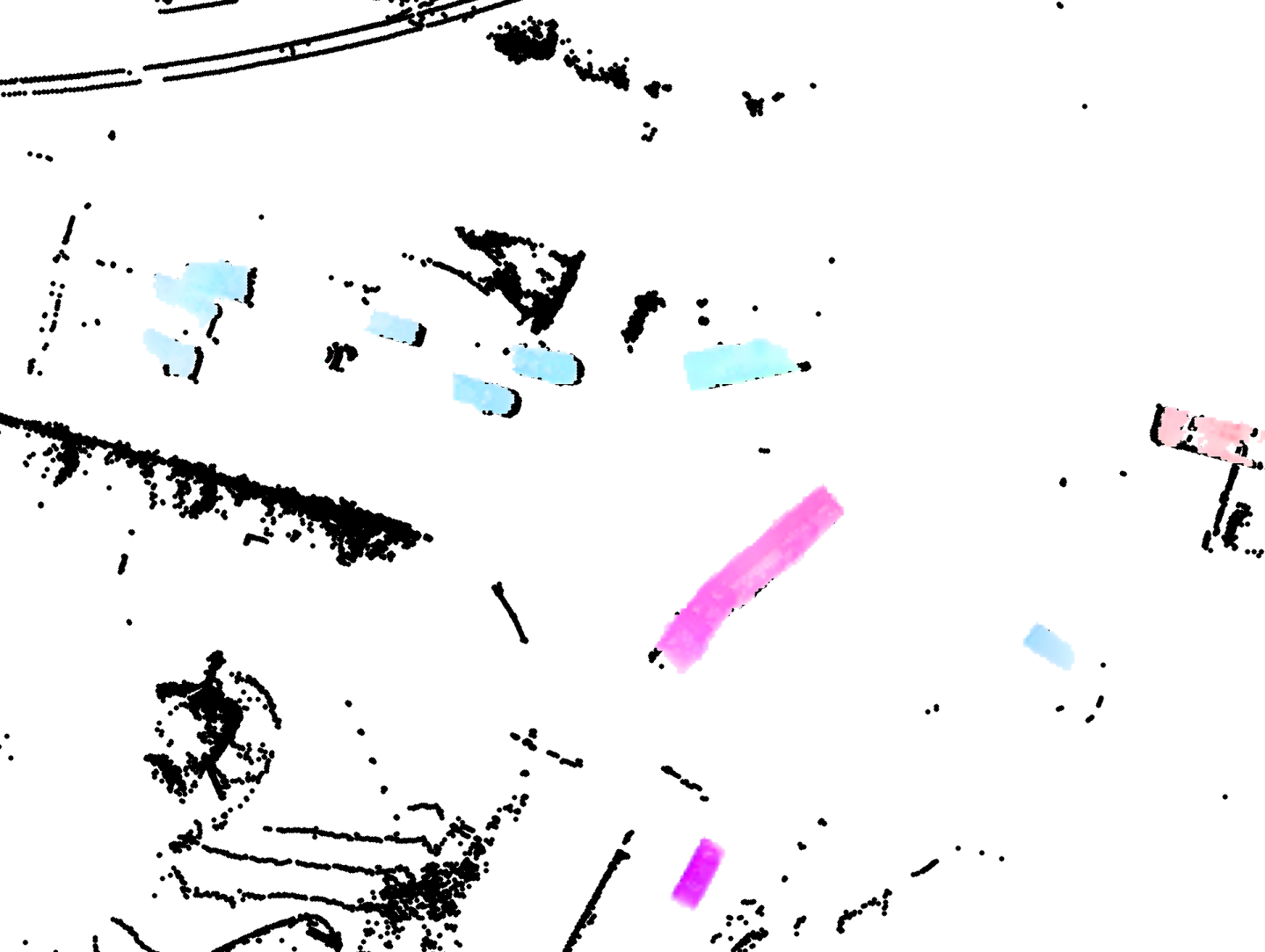}
\caption{An example of PillarFlow results shows that flows of vehicles are visualized with the colors indexing their directions~\cite{baker2011database}, where the long pink region in the figure is a truck towing a trailer. Notice that PillarFlow does not rely on object detection and classification results as inputs.}
\label{fig:flow_example}
\vspace{-6mm}
\end{figure}

Figure 1 shows an example of our flow estimation results in bird's eye view, where the moving vehicles are accurately inferred by the estimated flow (colorful regions).
Our \textbf{main contributions} can be summarized as follows.
\begin{itemize}
\item We propose an end-to-end method, \textit{PillarFlow}, to effectively estimate dense local 2-D motion in a LIDAR BeV representation.
Our deep net can leverage contextual knowledge of the scene and generalize to properly estimate the motion of unseen object types.

\item We integrate and leverage our proposed 2-D BeV flow estimation to improve object tracking performance on both a public and an internal dataset.
\item We demonstrate the computational efficiency, robustness, and practical use of our approach by integrating it in a real-world autonomous driving platform operating in challenging urban conditions.
\end{itemize}

The rest of the paper is organized as follows.
Section~\ref{sec:related_work} gives a brief overview of related work.
Section~\ref{sec:method} depicts the proposed system and network architectures.
We present our experimental results on public and in-house datasets in Section~\ref{sec:experiments}, followed by a real-world integration and in-depth analysis in an autonomous car in Section~\ref{sec:exp_autonomous}.
We report our conclusions in Section~\ref{sec:conclusions}.

\begin{figure*}[ht!]
\centering
\includegraphics[width=\textwidth]{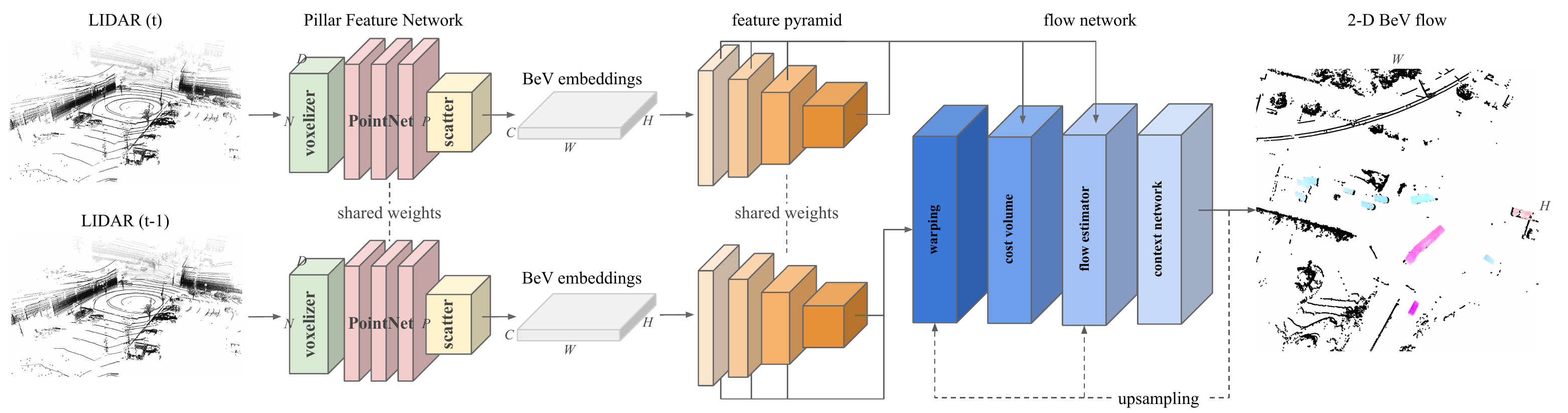}
\vspace{-6mm}
\caption{
\textbf{Proposed PillarFlow network architecture.} Two LIDAR sweeps are encoded by a Pillar Feature Network \cite{lang2019pointpillars} to obtain 2-D BeV embeddings, which are then fed into a feature pyramid network. The pyramid features are fed into a warping function, a cost volume layer, and a flow estimator to fuse the extracted features in the current scale level with the estimated flows from lower scales. Finally, a context network is used for flow refinement.}
\label{fig:overview}
\vspace{-4mm}
\end{figure*}

\section{RELATED WORK}
\label{sec:related_work}

%
%

\subsection{Scene Flow Estimation}
To estimate motion in the surrounding world, many approaches have been developed to estimate scene flow directly from LIDAR sweeps.
For instance, Dewan \etal~\cite{dewan2016rigid} formulate rigid scene flow estimation as an energy minimization problem using SHOT feature descriptors.
%
Ushani \etal~\cite{ushani_learning_2017} propose a learning-based flow estimation that trains an encoding network to extract binary vector features from 3-D points. Instead, we use pillar feature for better representation.
Other works~\cite{vaquero_deep_2018,baur_real-time_2019} rely on range images projected from LIDARs for flow estimation in projective view.
%
%

A common alternative for scene flow estimation is to use unstructured point-based representations.
Gu \etal~\cite{HPLFlowNet} propose an end-to-end deep network to fuse features from unstructured point clouds from two consecutive LIDAR sweeps.
Wang \etal~\cite{Wang_2018_CVPR} propose a parametric continuous convolution layer for non-grid-structured data, and demonstrate the application in point cloud segmentation and LiDAR motion estimation.
Liu \etal~\cite{liu2019flownet3d} introduce FlowNet3D, which builds on PointNet++~\cite{qi2017pointnet++}, leveraging a flow embedding layer to fuse two consecutive LIDAR sweeps.
Wang \etal~\cite{wang_flownet3d_2019} improve on FlowNet3D by using additional geometric loss functions beyond the L2 distance (Point to Plane and Cosine Distance).
Behl \etal~\cite{behl2019pointflownet} propose PointFlowNet to jointly train the tasks of 3-D scene flow, rigid motion prediction, and 3D object detection from unstructured LIDAR data.
Recently, self-supervised learning has also shown promise for 3-D scene flow estimation~\cite{mittal_just_2019,wu_pointpwc-net_2019}.
%
%
%

The aforementioned 3-D scene flow methods focus on accurately predicting point-to-point correspondences. %
They often suffer from high computational costs, which is a key challenge for real-time deployment on a robotic platform.

\subsection{Occupancy Grid Maps}

Occupancy grid maps (OGMs) are widely used to represent scene obstacle occupancy for robotics applications.
Ondruska \etal~\cite{ondruska2016deep} propose a deep tracking framework which incorporates a simple RNN to learn OGM-to-OGM mappings.
Ushani \etal~\cite{ushani_feature_2018} formulate 2-D BeV flow estimation as a similarity learning problem by transferring 3-D OGMs into 2-D embedding grids.
A separate classifier learns the matched foreground cells between frames by using an expectation maximization algorithm.
Later, Dequaire \etal~\cite{dequaire2018deep} extend the work of Ondruska \etal~\cite{ondruska2016deep} by using a spatial transformer module and dilated gated recurrent units  to account for observations from a moving platform.
%
%
Wirges \etal~\cite{wirges_self-supervised_2019} propose a learned approach to determine a motion mask on an OGM but uses hand crafted input features such as mean intensity and height range of points falling within each cell, rather than raw point clouds. 

Estimation of the per cell motion state within an occupancy grid is a recent topic referred to as dynamic occupancy grid maps (DOGMa) estimation. Online versions typically model this state using particle filtering.
Nuss \etal~\cite{nuss_random_2018} propose an efficient implementation of DOGMa using a particle filtering scheme, which we adopt later on in the paper for comparison with our proposed method.
Multiple methods have also been proposed to cluster and extract object level representations from a DOGMa for multi-object tracking~\cite{danescu_obstacle_2011,steyer_object_2017,vatavu_environment_2018,gies2018environment,steyer_grid-based_2019}.
Finally, some deep learning works build on the DOGMa representation for various tasks.
For instance, Hoermann \etal~\cite{hoermann2018dynamic} augment the DOGMa with a recurrent network trained by self-supervised labeling to predict future states.
Piewak \etal~\cite{piewak_fully_2017} build upon the Dynamic Occupancy Grid to do semantic segmentation of the DOGMa internal per cell state as static or dynamic.

While our method bears some similarities to the aforementioned works leveraging grid-based representation, we explore a new architecture bringing together both end-to-end flow techniques and grid-based representations.

\section{PROPOSED SYSTEM}
\label{sec:method}

We propose a method, \textit{PillarFlow}, to learn to estimate 2-D BeV flow by combining a Pillar Feature Network (PFN)~\cite{lang2019pointpillars} with a flow estimation network.
The overview of the system is shown in Figure~\ref{fig:overview}.
First, two consecutive point cloud sweeps are aligned into the same coordinate frame: the original coordinates of the LIDAR sweep at $t-1$ are transformed to the coordinate frame of the LIDAR sweep at $t$ using the odometry information of the robot.
Next, the two point clouds are encoded by PFN to build two BeV pseudo-images where each cell has a learned embedding based on points that fall inside it.
Then the two pseudo images are fed to a flow network to estimate the dense 2-D in the BeV space.

\subsection{3-D Point Cloud to 2-D BeV Embedding}

In our system, we use a PFN to extract 2-D BeV embeddings from 3-D point clouds. 
First, a voxelization step is applied to the point cloud by discretizing the $x$-$y$ plane, thus creating a set of ``pillars'' (grid cells) in birds-eye-view.
The voxelized pointcloud is structured as a $(D, P, N)$-shaped tensor where $D$ is the number of point descriptors, $P$ is the number of pillars, and $N$ is the number of points per pillar.
We use $D = 9$, where the first four values denote coordinates $x, y, z$ and reflectance $r$.
The next five values are the distances to the arithmetic mean $x_c, y_c, z_c$ of all points in a pillar and the offset $x_p, y_p$ from the pillar center.
%
Next, this input tensor is processed by a simplified version of PointNet~\cite{pointnet} to get a feature map of shape $(C, P, N)$. We further compress the feature map by a max operation over the last dimension, resulting in a $(C,P)$ encoded feature map with a $C$-dimensional feature embedding for each pillar.
Finally, the encoded features are scattered back to original pillar locations to create a pseudo-image tensor of shape $(C, H, W)$, where $H$ and $W$ indicate the height and width of the pseudo-image.

\subsection{2-D BeV Flow Network}

To accurately associate the embeddings, i.e., pillar features between 2-D BeV grids, we conduct a 2-D BeV flow estimation.
%
Based on the PWC-Net model~\cite{sun2018pwc}, we adjust architecture parameters such as receptive field and correlation layer parameters to account for the maximum relative motion that would be expected to be encountered between consecutive LIDAR sweeps (given the time delta between frames, grid resolution, and typical vehicle speeds).

As depicted in Figure~\ref{fig:overview}, the pillar features are further encoded via a feature pyramid network.
A cost volume layer is then used to estimate the flow, where the matching cost is defined as the correlation between the two feature maps.
Finally, a context network is applied to exploit contextual information for additional refinement.
The context network is a feed-forward CNN based on dilated convolutions, along with batch normalization and ReLU .
%

\subsection{Training}
We use annotated object tracks to generate 2-D BeV flow ground truth.
For each object, we estimate the instantaneous velocity from the difference in object positions divided by the elapsed time between consecutive frames.
Our assumption is that only labeled dynamic objects can have a valid velocity, and all non-labeled obstacles and background should have zero velocity.
Note that this assumption might be violated in practice and does not provide direct supervision for potential out-of-ontology moving objects (static objects have zero flow). Nonetheless, our experiments show that this provides enough supervision to learn an object-agnostic, dense (pillar-level) optical flow network on BeV grids that generalizes well. Another exciting possibility we leave for future work would be to extend self-supervised approaches like~\cite{wu_pointpwc-net_2019}.
%

Let $\mathbf{\hat{f}}^l_{\theta}$ denote the flow field at the $l^{\text{th}}$ pyramid level predicted by the network with learnable parameters $\theta$, and $\mathbf{f}^l_{gt}$ the corresponding ground truth.
We use the common multi-scale training loss from~\cite{ilg2017flownet} and~\cite{sun2018pwc}:

\begin{equation}
\mathcal{L} = \sum_{l=l_0}^{L} \alpha_l\sum_{x}|\mathbf{\hat{f}}^l_{\theta} - \mathbf{f}^l_{gt}|_2,
\label{eq:training_loss}
\end{equation}

\noindent where $|\cdot|_2$ is the L2 norm and the weights $\alpha_l$ in the training loss are set by following the setting from Sun \etal~\cite{sun2018pwc}.


\section{Experimental Results}
\label{sec:experiments}

We conduct a thorough evaluation and analysis at the 2-D BeV flow estimation, and at the system-level by integrating our approach into different tracking systems.

\subsection{Datasets}

To evaluate the performance of the proposed method, we conduct experiments on two datasets. 
First, we use \textit{nuScenes} \cite{nuscenes2019}, a public large-scale dataset for autonomous driving development, containing multiple sensors data such as LIDAR, radar, and cameras.
The dataset includes 850 scenes for training and 150 scenes for validation, with fully annotated detection and tracking labels.
Second, we use an in-house dataset, \textit{TRI-cuboid}, collected by our fleet of autonomous cars equipped with LIDAR, radar, and several cameras.
The dataset includes 194 scenes for training and 40 scenes for validation with fully annotated 3D bounding boxes for a variety of object categories.

\subsection{Implementation details}

We limit the range of the point cloud to $[-50, 50]$ meters in both $x$ and $y$ directions with a grid resolution of $0.25$ meter per cell.
During training, we use the Adam optimizer with exponentially decayed learning rate starting from $0.0001$ and then reduce it by a factor of $0.9$ at each $1/10$ of total iterations.
We train for 2M iterations on nuScenes, and 4M iterations on TRI-cuboid dataset.
For data augmentation, we apply (ego-centric) random rotation and random bounding box scaling to the LIDAR sweeps.
We perform simple ground plane removal to filter out points lying on a ground plane obtained with the RANSAC algorithm.

\subsection{2-D BeV Flow Estimation}

%
%
To evaluate the performance of our PillarFlow model, we compare it to two baselines for BeV flow estimation.

\textbf{Iterative Closest Point (ICP)}:
ICP~\cite{besl1992method} outputs a transformation associating two point clouds by using an SVD-based point-to-point algorithm.
Here we select the 3-D points within the clusters, and then apply ICP to obtain the transform between the two point clouds. The flow is inferred by the difference between the corresponding coordinates after the transformation. 

\textbf{Binary OGM}: Binary OGM binarizes the occupancy grid map from LIDAR sweeps.
Instead of pillar features, we use the binary OGMs as the inputs of the adapted one-channel PWCNet, which learns to estimate the 2-D flow in BeV grids.


\begin{table}[tp]
\centering
\resizebox{\columnwidth}{!}{
\begin{tabular}{|c||c|c|c|c|}
\hline
Dataset & \multicolumn{4}{c|}{nuScenes} \\
\hline
\diagbox[font=\scriptsize]{Method}
{Metric} & RMSE (dynamic) & RMSE (static) & RMSE (average) & AAE \\
\hline
ICP + Det. & 2.818 & NA & 2.818 & 0.216 \\
Binary OGM & 1.316 & 0.113 & 0.247 & \textbf{0.091} \\
\hline
Ours & \textbf{1.127} & \textbf{0.110} & \textbf{0.207} & 0.108 \\
\hline
\hline
Dataset & \multicolumn{4}{c|}{TRI-cuboid} \\
\hline
\diagbox{\scriptsize Method}
{\scriptsize Metric} & RMSE (dynamic) & RMSE (static) & RMSE (average) & AAE \\
\hline
ICP + Det. & 0.757 & NA & 0.757 & 0.681 \\
Binary OGM & 0.409 & 0.036 & 0.081 & 0.098 \\
\hline
Ours & \textbf{0.288} & \textbf{0.027} & \textbf{0.029} & \textbf{0.087} \\
\hline
\end{tabular}
}
\caption{Quantitative results in comparison to baselines on the nuScenes and TRI-cuboid datasets.}
\label{tab:baselines}
\vspace{-6mm}
\end{table}


\begin{table}[tp]
\centering
\begin{tabular}{|c||c|c|c|c|c|}
\hline
Dataset & \multicolumn{2}{c|}{nuScenes} & \multicolumn{2}{c|}{TRI-cuboid} \\
\hline
Flow Network Used & RMSE & AAE & RMSE & AAE \\
\hline 
SpyNet & 0.327 & 0.122 & 1.181 & 0.146 \\
FlowNet2 & 0.320 & \textbf{0.072} & 0.093 & 0.100 \\
PWCNet* (Ours) & \textbf{0.207} & 0.087 & \textbf{0.029} & \textbf{0.087} \\
\hline
\hline
Ground Plane Config & RMSE & AAE & RMSE & AAE \\
\hline
Ours w/ ground & 0.246 & 0.079 & 0.069 & 0.092 \\
Ours w/o ground &\textbf{0.207} & 0.087 & \textbf{0.029} & \textbf{0.087} \\
\hline
\end{tabular}
\caption{Ablation study of different flow network architectures.}
\label{tab:model_ablation}
\vspace{-8mm}
\end{table}

\textbf{Metrics:}
We use root mean square error (RMSE) in meter and average angular error (AAE) in radians as our metrics to measure flow error in 2-D BeV grids.

\begin{figure*}
    \centering
    \includegraphics[width=\textwidth]{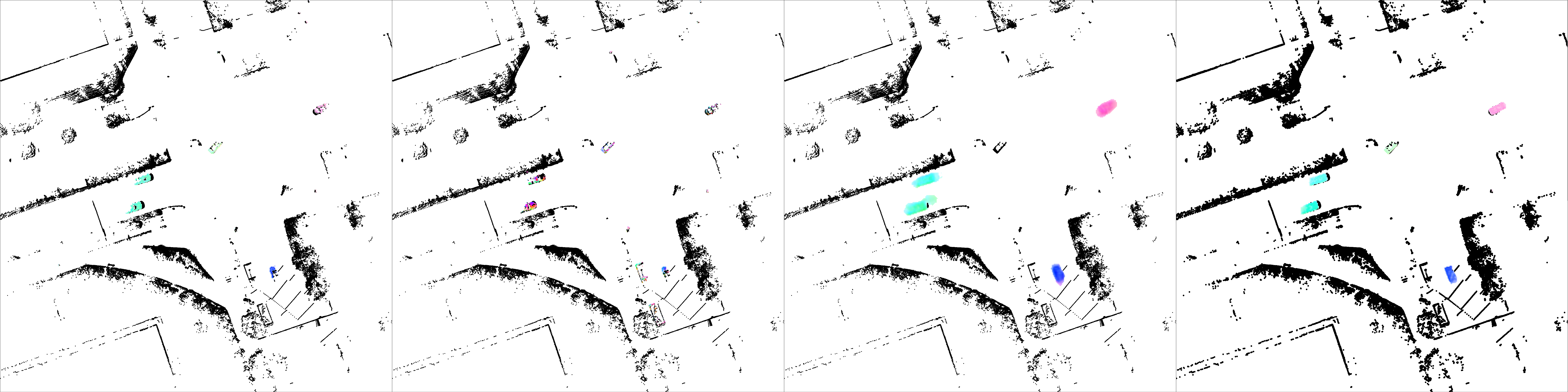}
    \\
    \includegraphics[width=\textwidth]{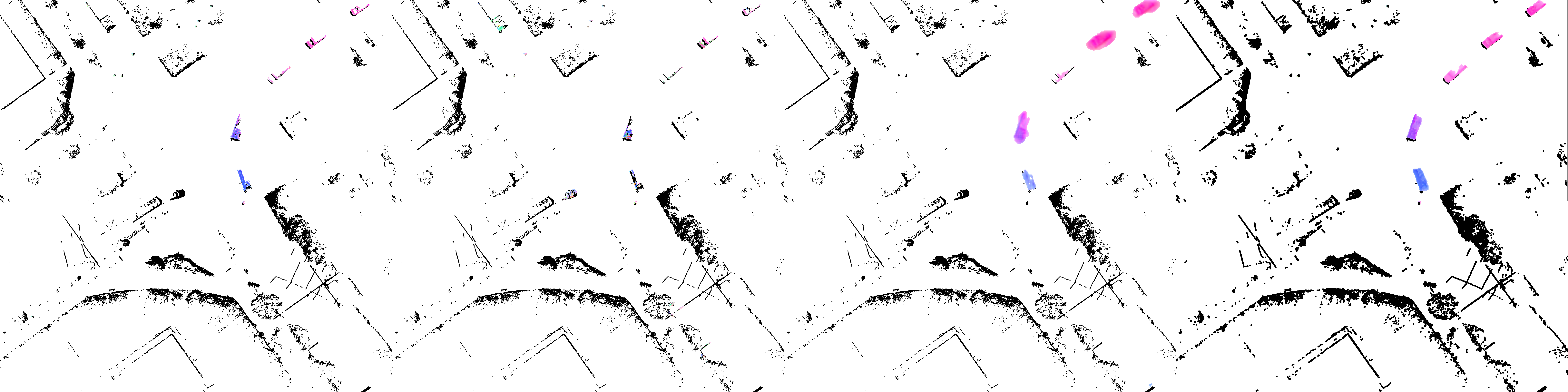}
    \vspace{-5mm}
    \caption{\textbf{Qualitative results of 2-D BeV grid flow.} Two rows present two examples in TRI-cuboid, where the figures from left to right are: groundtruth, IPC + Det., binary OGM, PillarFlow. The estimated 2-D BeV flows is visualized by an optical flow representation in \cite{baker2011database}.} \label{fig_autonomous_example}
    \label{fig:flow_qualitative}
    \vspace{-2mm}
\end{figure*}

\textbf{Results:}
Table~\ref{tab:baselines} shows the quantitative results comparing PillarFlow to the baselines, where the dynamic objects indicate movable objects such as vehicle, pedestrian, bicycle, truck, etc, and the static ones are immovable like building, pole, vegetation, etc.
Our proposed method achieves an average error of $2m$ in nuScenes and $0.3 m$ in TRI-cuboid.
Compared to the baselines, our method can better handle the flow produced by dynamic objects.
Figure \ref{fig:flow_qualitative} shows qualitative results on the TRI-cuboid dataset, confirming that the 2-D BeV flows are accurately estimated.
We observe that our method can deal better with empty cells and dynamic objects observed as stationary for which
 we predict zero flow while the baselines display significant noise on those objects.

To better analyze different components of our proposed approach, we compare against alternative flow network architectures, i.e., SpyNet \cite{ranjan2017optical} and FlowNet2 \cite{ilg2017flownet}. 
The comparison results are reported in Table~\ref{tab:model_ablation}.
The result indicates that PWCNet (the adaptation is mentioned in Section~\ref{sec:method}) is the most compatible with the proposed system, showing a better capability of handling not only dynamic objects but also static ones.
We also conduct another analysis of the impact of ground point removal, which yields an improvement of only $0.4m$ RMSE on both nuScenes and TRI-cuboid datasets.
This also implies that the proposed method can perform properly even without filtering the ground plane.

\begin{table}[tp]
\centering
\resizebox{\columnwidth}{!}{
    \begin{tabular}{|c||c|c||c|c|}
    \hline
    Dataset & \multicolumn{4}{c|}{nuScenes} \\
    \hline
    \diagbox[innerwidth=4cm]{Method}{Metric} & AMOTA & AMOTP (m) & MOTA & MOTP (m) \\ \hline \hline
    StanfordIPRL-TRI \cite{chiu2020probabilistic} & 56.1 & \textbf{80.0} & 48.3 & 37.9 \\
    \hline
    StanfordIPRL-TRI + Binary OGM  & 56.5  & 79.6  & {48.6} & \textbf{38.2} \\
    \hline
    StanfordIPRL-TRI + Ours & \textbf{56.6} &79.7  & \textbf{49.2} & \textbf{38.2} \\
    \hline
    \end{tabular}
}
\caption{
Performance on nuScene Tracking Validation Scenes.}
\label{tab:tracking_quantitative}
\vspace{-8mm}
\end{table}

\subsection{Tracking Evaluation on nuScenes} 
In this section, we provide an evaluation of our proposed 2-D BeV flow estimation approach integrated in a state-of-the-art tracking framework on nuScenes tracking dataset.
We adopt the tracking pipeline of the winner entry to nuScenes Tracking Challenge at NeurIPS 2019~\cite{nuscene-tracking}, StanfordIPRL-TRI~\cite{chiu2020probabilistic} as our baseline.
The baseline takes 3-D object detection results as measurement source and parameterizes them into a 7 dimension observation state: $\textbf{o}_t = (x,y,z,a,l,w,h)^T $.
To integrate our proposed 2-D BeV flow estimation, we extend the observation state with 2-D object velocity, $(d_x, d_y)$.
The object level velocity is approximated by the mean flow vector of all the BeV grids within the detected bounding box boundary.
Under independent assumption between object detections and velocity estimations, our new observation model and covariance matrix are given as follows:
\begin{equation}
\hat{\textbf{o}}_t = [\textbf{o}_t^T,d_x,d_y]^T,
\end{equation}
\begin{equation}
       \hat{\textbf{R}} =
\begin{bmatrix}
\textbf{R}   & \textbf{0}\\ 
 \textbf{0}  & \textbf{R}_d\\
\end{bmatrix} ,
\end{equation}
where $\textbf{R}$ is the original noise covariance and $\textbf{R}_d$ is the noise covariance matrix for velocity.
Following the suggestion by Chiu \etal~\cite{chiu2020probabilistic}, we estimate $\textbf{R}_d$ from the training set.

We report the tracking performance on the nuScenes tracking validation scenes in Table~\ref{tab:tracking_quantitative}.
We use the same evaluation metrics suggested by the nuScenes Tracking Challenge~\cite{nuscene-tracking}, including average multi-object tracking accuracy (AMOTA), average multi-object tracking precision (AMOTP), multi-object tracking accuracy (MOTA), and multi-object tracking precision (MOTP).
We compare our proposed approach to the baseline that did not use velocity in observation. We also compare to different velocity estimations provided by other BeV flow baselines (discussed in the Section~\ref{sec:experiments}-B) in the same tracking algorithm.
The results indicate that our estimated velocity is able to further improve the state-of-the-art tracking performance working in parallel with 3D object detections.
Compared with the velocity generated by binary OGM, our approach achieves better tracking stability (AMOTA) with less degradation in positional precision (AMOTP), which is a commonly seen trade-off in multi-object tracking~\cite{nuscene-tracking}.
The better trade-off also reflects the accuracy and robustness of our proposed 2-D BeV flow estimation method.


\section{Systematic Analysis on Autonomous Vehicle}
\label{sec:exp_autonomous}

In this section, we provide a full-system integration of a class-agnostic tracking system and further discuss the advantage of our proposed method that is not fully captured in the previous metrics.
We integrate the proposed method into an in-house autonomous platform, and quantitatively evaluate it on 43 different 10 second snippet scenes collected from the Odaiba area of Tokyo, Japan, and Ann Arbor, Michigan, US.
These logs have been fully annotated with bounding cuboids for ontology based objects.
In this section, we briefly describe our tracking pipeline, and analyze how the proposed method improves the tracking performance of the generic objects.

\subsection{Object Tracker Pipeline}

To detect generic object clusters, we use a  2.5-D terrain height map to filter out non-object LIDAR points, by eliminating those falling outside the range of [$0.3$m, $2.0$m] above the corresponding ground cell height.
Then, we collapse the remaining points to a 2D occupancy grid and utilize a connected-components clustering algorithm to cluster the remaining points into class-agnostic objects.
We use a dilated object mask derived from a LIDAR object detector to enforce constraints on the labeling to ensure that the clusters are correctly segmented. 
Such a cluster representation can achieve a high object recall and ensure that all measurements are accounted for, which is important for safety, especially if a detection is not triggered.

The observation state of an object track is represented as $(x, y, v_x, v_y, a_x, a_y)^T$.
We apply a nearest-neighbor association of existing tracks to object measurements by computing the Mahalanobis distance between the track's predicted position and LIDAR object cluster's position.
A gating threshold on the nearest result is used to either construct an association, if close enough, or create a new track otherwise.
%
To estimate the state of each object track, we use a fixed-lag smoothing approach built upon a factor graph representation \cite{dellaert_factor_2017}. %
Binary factors between consecutive object state nodes encode a constant acceleration motion model.
These factors attempt to minimize the difference in acceleration between consecutive states, with the error residuals weighed according to a predetermined square root information matrix.
If a measurement (LIDAR object cluster) is associated to a track, a new state node is created and linked with the existing graph using this motion model prior.
A unary prior factor is added to the new state node to represent the measurement, minimizing the difference between the measurement's position and the corresponding state node's position.



\begin{table*}[tp]
\centering
\resizebox{\textwidth}{!}{
\begin{tabular}{|c||c|c|c||c|c|c|}
\hline
Metrics & \multicolumn{3}{c||}{Mean Track Velocity Error in m/s} & \multicolumn{3}{c|}{95\% Largest Track Velocity Error in m/s} \\
\hline
\diagbox[innerwidth=4cm]{Category}{Methods} & Baseline & Baseline + DOGMa & Baseline + PF & Baseline & Baseline + DOGMa & Baseline + PF \\
\hline
Static Objects & 0.839 & 0.848 & \textbf{0.480} & 3.993 & 3.803 & \textbf{2.322} \\
\hline
Pedestrian \& Cyclist &  0.772 & \textbf{0.523} & 0.641 & 3.411 & 1.621 & \textbf{1.446} \\
\hline
Objects observed stationary & 0.861 & 0.512 & \textbf{0.059} & 3.826 & 1.796 & \textbf{0.151} \\
\hline
Slow Moving Objects $(0, 3]$ m/s & \textbf{0.566} & 0.570 & 0.666 & 2.117 & 1.709 & \textbf{1.560} \\
\hline
Fast Moving Objects $[3, \infty)$ m/s & 2.396 & 2.371 & \textbf{2.036} & 15.188 & 11.490 & \textbf{7.468} \\
\hline
\end{tabular}
}
\caption{Comparison of Mean Track Velocity Error in m/s.}

\label{table:tracking_95}
\vspace{-1mm}
\end{table*}

\begin{figure*}
\centering
\includegraphics[width=\columnwidth]{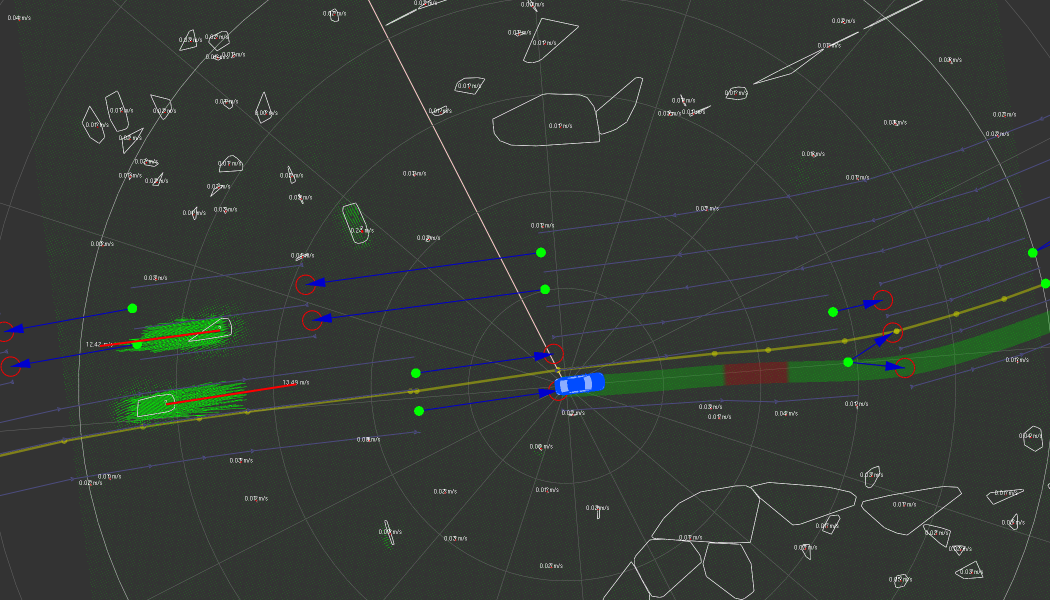}
\includegraphics[width=\columnwidth]{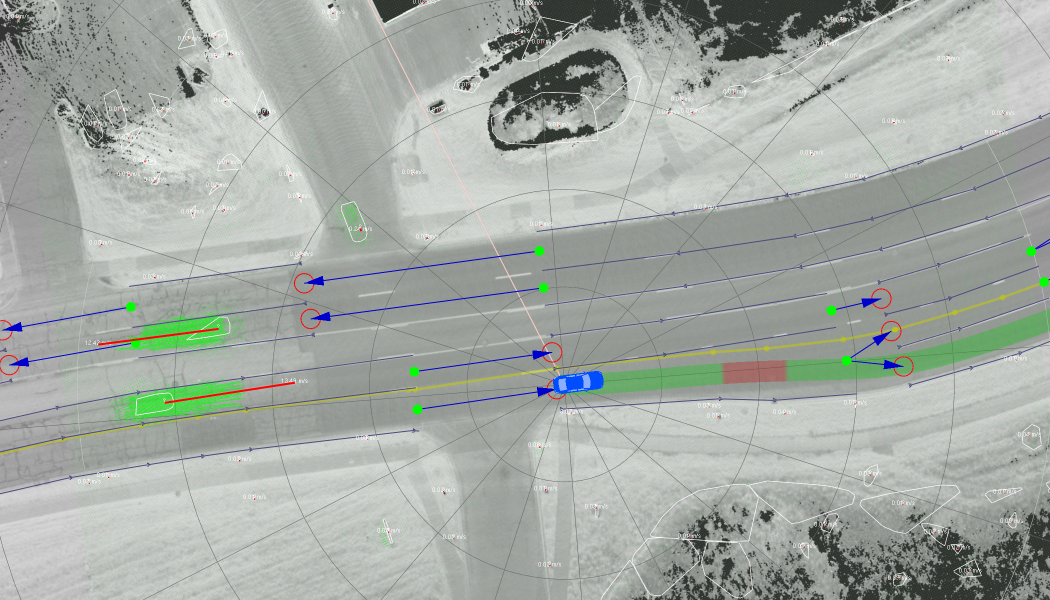}
\vspace{-1mm}
\caption{Tracker Integration Result. The green lines represent the flow vectors for each grid cell. The red lines represent the aggregated velocity for each object. This image demonstrates how static environmental objects do not get spurious velocities despite being observed from a moving platform, once our method is applied. }
\label{fig:bev_flow}
\end{figure*}


\begin{figure*}
\centering
\includegraphics[width=\columnwidth]{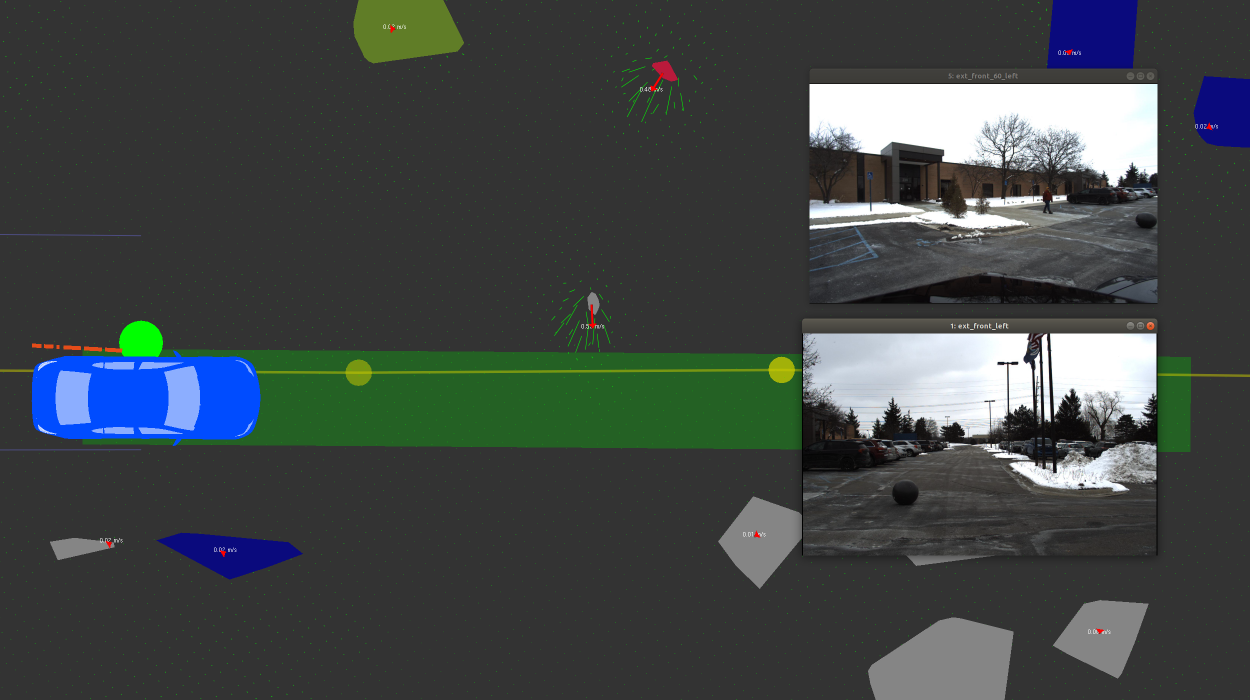}
\includegraphics[width=\columnwidth]{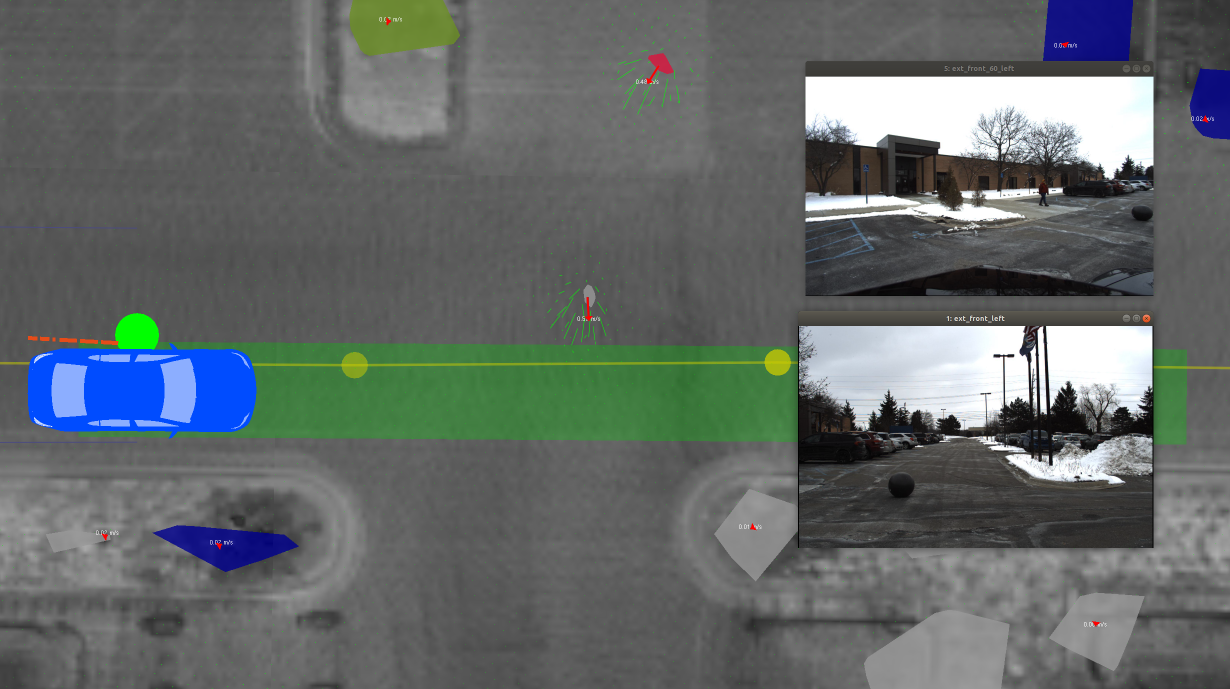}
\vspace{-1mm}
\caption{We conduct a qualitative evaluation on an out-of-ontology generic moving object. We throw a medicine ball and note that the proposed architecture has sufficiently generalized to estimate motion from this unseen object types. }
\label{fig:generic_flow}
\vspace{-2mm}
\end{figure*}

\subsection{Experimental Settings}


We augment the tracker described above by adding an additional measurement prior, object velocity measurement, estimated from our proposed method (\textit{PillarFlow}). We compare our estimated velocity with another non-learning based velocity estimation methods widely adapted in real-world system, DOGMa~\cite{nuss_random_2018}.
%
%
Integration details adapting the two methods are given below:

\textbf{PillarFlow} (PF):
We aggregate the 2-D BeV flows as estimated by our method to compute a single mean velocity and co-variance per object cluster.
This is simply obtained by random sampling the set of 2-D BeV flows from the cells occupied by the object cluster.

\textbf{DOGMa}: 
DOGMa models the dynamic state estimation of grid cells as a random finite set (RFS) problem.
We implement the DOGMa approach~\cite{nuss_random_2018} for comparison.
Similarly, we aggregate the motion vectors to a mean per cluster by sampling, and weighing each sample based on the occupancy probability of the cell.
%
Moreover, a weighted aggregation of the DOGMa's preexisting velocity co-variance matrix per occupied cell is applied to propagate the uncertainty of all cells.
%
Note that this filtered approach can contain more information than just a two frame instantaneous estimate.



To compare the approaches, we evaluate the track versus ground truth object velocity error, instead of the typical tracking metrics such as MOTA and MOTP, false positive alarm rate, or track ID switch count.
The reason is that generic objects are unlabeled and thus wouldn't be reflected in the typical metrics. 
To associate a track with a ground truth annotated object for evaluation purposes, we require a significant overlap between the track's most recent associated object cluster's convex hull and a ground truth bounding box at the object cluster's timestamp.
Due to lack of bounding boxes for generic objects such as guardrails or trees, we assume that tracks that fail the association represent a generic LIDAR object with zero velocity.

\subsection{Analysis and Discussion}

The quantitative results in Table ~\ref{table:tracking_95} show strong enhancements to tracking performance using our proposed 2-D BeV flow estimation as a prior.
Overall, mean and worst case performance are improved across most object class types.
In particular, the proposed method improves significantly in stationary objects (e.g., parked cars, standing pedestrians, and objects excluding static background).

We observe that the prior velocity information provides faster initialization and convergence of the track's state estimate and allow for better data association, thereby creating more consistent tracks.

We provide qualitative results of the tracker integration in out autonomous platform, and an out-of-ontology tracker observation in Figures~\ref{fig:bev_flow} and~\ref{fig:generic_flow} respectively.

%
%

\begin{table}[tp]
\centering
\begin{tabular}{|c||c|c||c|}
\hline
Methods & ICP + Det. & DOGMa & PillarFlow \\ \hline \hline
mean (ms) & 20 & 45 & 23 \\
\hline
variance (ms) & 5 & 1 & 0.1 \\
\hline
\end{tabular}
\caption{Runtime Performance on the in-house  Autonomous Platform}
\label{tab:runtime_table}
\vspace{-8mm}
\end{table}

\subsection{Runtime Performance}

Table~\ref{tab:runtime_table} summarizes the runtime performance onboard our autonomous test platform.
Our proposed model can achieve approximately 40Hz when run via TensorRT with half-precision floating point mode on a single Nvidia Quadro RTX 6000 GPU.
This demonstrates that the proposed method is feasible for real-time accurate velocity estimation in real-world applications.

\section{CONCLUSION}
\label{sec:conclusions}

We propose \textit{PillarFlow}, a deep network for end-to-end dense motion estimation on 2-D BeV grids. 
Experimental results show that \textit{PillarFlow} improves the performance of dynamic object tracking on two datasets.
Additionally, we demonstrate that \textit{PillarFlow} delivers substantial improvements for real-time generic obstacle tracking onboard a real-world autonomous car.
%
%
Nonetheless, we notice occasional incorrect flow predictions due to (dis)occlusions or for objects with few points.
Interesting future work includes
accumulating more temporal context or separately estimating occlusions and then augmenting the network input~\cite{ferrari_occlusions_2018}.






\begin{thebibliography}{10}
\providecommand{\url}[1]{#1}
\csname url@rmstyle\endcsname
\providecommand{\newblock}{\relax}
\providecommand{\bibinfo}[2]{#2}
\providecommand\BIBentrySTDinterwordspacing{\spaceskip=0pt\relax}
\providecommand\BIBentryALTinterwordstretchfactor{4}
\providecommand\BIBentryALTinterwordspacing{\spaceskip=\fontdimen2\font plus
\BIBentryALTinterwordstretchfactor\fontdimen3\font minus
  \fontdimen4\font\relax}
\providecommand\BIBforeignlanguage[2]{{%
\expandafter\ifx\csname l@#1\endcsname\relax
\typeout{** WARNING: IEEEtran.bst: No hyphenation pattern has been}%
\typeout{** loaded for the language `#1'. Using the pattern for}%
\typeout{** the default language instead.}%
\else
\language=\csname l@#1\endcsname
\fi
#2}}

\bibitem{badue2019self}
C.~Badue, R.~Guidolini, R.~V. Carneiro, P.~Azevedo, V.~B. Cardoso,
  \emph{et~al.}, ``Self-driving cars: A survey,'' \emph{arXiv:1901.04407},
  2019.

\bibitem{chen2017multi}
X.~Chen, H.~Ma, J.~Wan, B.~Li, and T.~Xia, ``Multi-view {3D} object detection
  network for autonomous driving,'' in \emph{CVPR}, 2017.

\bibitem{ku2018joint}
J.~Ku, M.~Mozifian, J.~Lee, A.~Harakeh, and S.~L. Waslander, ``Joint {3D}
  proposal generation and object detection from view aggregation,'' in
  \emph{IROS}, 2018.

\bibitem{yang2018pixor}
B.~Yang, W.~Luo, and R.~Urtasun, ``Pixor: Real-time {3D} object detection from
  point clouds,'' in \emph{CVPR}, 2018.

\bibitem{zhou2018voxelnet}
Y.~Zhou and O.~Tuzel, ``{VoxelNet}: End-to-end learning for point cloud based
  {3D} object detection,'' in \emph{CVPR}, 2018.

\bibitem{lang2019pointpillars}
A.~H. Lang, S.~Vora, H.~Caesar, L.~Zhou, J.~Yang, and O.~Beijbom,
  ``{PointPillars}: Fast encoders for object detection from point clouds,'' in
  \emph{CVPR}, 2019.

\bibitem{held2014combining}
D.~Held, J.~Levinson, S.~Thrun, and S.~Savarese, ``Combining {3D} shape, color,
  and motion for robust anytime tracking.'' in \emph{RSS}, 2014.

\bibitem{weng2019baseline}
X.~Weng and K.~Kitani, ``A baseline for {3D} multi-object tracking,''
  \emph{arXiv:1907.03961}, 2019.

\bibitem{chiu2020probabilistic}
H.-k. Chiu, A.~Prioletti, J.~Li, and J.~Bohg, ``Probabilistic {3D} multi-object
  tracking for autonomous driving,'' \emph{arXiv:2001.05673}, 2020.

\bibitem{buehler2020driving}
A.~Buehler, A.~Gaidon, A.~Cramariuc, R.~Ambrus, G.~Rosman, and W.~Burgard,
  ``Driving through ghosts: Behavioral cloning with false positives,'' in
  \emph{IROS}, 2020.

\bibitem{sun2018pwc}
D.~Sun, X.~Yang, M.-Y. Liu, and J.~Kautz, ``{PWC-Net}: {CNNs} for optical flow
  using pyramid, warping, and cost volume,'' in \emph{CVPR}, 2018.

\bibitem{baker2011database}
S.~Baker, D.~Scharstein, J.~Lewis, S.~Roth, M.~J. Black, and R.~Szeliski, ``A
  database and evaluation methodology for optical flow,'' \emph{IJCV}, vol.~92,
  no.~1, pp. 1--31, Nov. 2011.

\bibitem{dewan2016rigid}
A.~Dewan, T.~Caselitz, G.~D. Tipaldi, and W.~Burgard, ``Rigid scene flow for
  {3D} lidar scans,'' in \emph{IROS}, 2016.

\bibitem{ushani_learning_2017}
A.~K. Ushani, R.~W. Wolcott, J.~M. Walls, and R.~M. Eustice, ``A learning
  approach for real-time temporal scene flow estimation from {LIDAR} data,'' in
  \emph{ICRA}, 2017.

\bibitem{vaquero_deep_2018}
V.~Vaquero, A.~Sanfeliu, and F.~Moreno-Noguer, ``Deep {Lidar} {CNN} to
  understand the dynamics of moving vehicles,'' in \emph{ICRA}, 2018.

\bibitem{baur_real-time_2019}
S.~A. Baur, F.~Moosmann, S.~Wirges, and C.~B. Rist, ``Real-time {3D} {LiDAR}
  flow for autonomous vehicles,'' in \emph{IV}, 2019.

\bibitem{HPLFlowNet}
X.~Gu, Y.~Wang, C.~Wu, Y.~J. Lee, and P.~Wang, ``{HPLFlowNet}: {Hierarchical
  Permutohedral Lattice FlowNet} for scene flow estimation on large-scale point
  clouds,'' in \emph{CVPR}, 2019.

\bibitem{Wang_2018_CVPR}
S.~Wang, S.~Suo, W.-C. Ma, A.~Pokrovsky, and R.~Urtasun, ``Deep parametric
  continuous convolutional neural networks,'' in \emph{CVPR}, 2018.

\bibitem{liu2019flownet3d}
X.~Liu, C.~R. Qi, and L.~J. Guibas, ``{FlowNet3D}: Learning scene flow in {3D}
  point clouds,'' in \emph{CVPR}, 2019.

\bibitem{qi2017pointnet++}
C.~R. Qi, L.~Yi, H.~Su, and L.~J. Guibas, ``{PointNet++}: Deep hierarchical
  feature learning on point sets in a metric space,'' in \emph{NeurIPS}, 2017.

\bibitem{wang_flownet3d_2019}
Z.~Wang, S.~Li, H.~Howard-Jenkins, V.~A. Prisacariu, and M.~Chen,
  ``{FlowNet3D++}: Geometric losses for deep scene flow estimation,''
  \emph{arXiv:1912.01438}, 2019.

\bibitem{behl2019pointflownet}
A.~Behl, D.~Paschalidou, S.~Donn{\'e}, and A.~Geiger, ``{PointFlowNet}:
  Learning representations for rigid motion estimation from point clouds,'' in
  \emph{CVPR}, 2019.

\bibitem{mittal_just_2019}
H.~Mittal, B.~Okorn, and D.~Held, ``Just go with the flow: Self-supervised
  scene flow estimation,'' \emph{arXiv:1912.00497}, 2019.

\bibitem{wu_pointpwc-net_2019}
W.~Wu, Z.~Wang, Z.~Li, W.~Liu, and L.~Fuxin, ``{PointPWC-Net}: {A}
  coarse-to-fine network for supervised and self-supervised scene flow
  estimation on {3D} point clouds,'' \emph{arXiv:1911.12408}, 2019.

\bibitem{ondruska2016deep}
P.~Ondruska and I.~Posner, ``Deep tracking: Seeing beyond seeing using
  recurrent neural networks,'' in \emph{AAAI}, 2016.

\bibitem{ushani_feature_2018}
A.~K. Ushani and R.~M. Eustice, ``Feature learning for scene flow estimation
  from {LIDAR},'' in \emph{CoRL}, 2018.

\bibitem{dequaire2018deep}
J.~Dequaire, P.~Ondr{\'u}{\v{s}}ka, D.~Rao, D.~Wang, and I.~Posner, ``Deep
  tracking in the wild: End-to-end tracking using recurrent neural networks,''
  \emph{IJRR}, vol.~37, no. 4-5, pp. 492--512, 2018.

\bibitem{wirges_self-supervised_2019}
S.~Wirges, J.~Gräter, Q.~Zhang, and C.~Stiller, ``Self-supervised flow
  estimation using geometric regularization with applications to camera image
  and grid map sequences,'' \emph{arXiv:1904.12599}, Apr. 2019.

\bibitem{nuss_random_2018}
D.~Nuss, S.~Reuter, M.~Thom, T.~Yuan, G.~Krehl, M.~Maile, A.~Gern, and
  K.~Dietmayer, ``A random finite set approach for dynamic occupancy grid maps
  with real-time application,'' \emph{IJRR}, vol.~37, no.~8, pp. 841--866, July
  2018.

\bibitem{danescu_obstacle_2011}
R.~G. Danescu, ``Obstacle detection using dynamic particle-based occupancy
  grids,'' in \emph{DICTA}, 2011.

\bibitem{steyer_object_2017}
S.~Steyer, G.~Tanzmeister, and D.~Wollherr, ``Object tracking based on
  evidential dynamic occupancy grids in urban environments,'' in \emph{IV},
  2017.

\bibitem{vatavu_environment_2018}
A.~Vatavu, N.~Rexin, S.~Appel, T.~Berling, S.~Govindachar, \emph{et~al.},
  ``Environment estimation with dynamic grid maps and self-localizing
  tracklets,'' in \emph{ITSC}, 2018.

\bibitem{gies2018environment}
F.~Gies, A.~Danzer, and K.~Dietmayer, ``Environment perception framework fusing
  multi-object tracking, dynamic occupancy grid maps and digital maps,'' in
  \emph{ITSC}, 2018.

\bibitem{steyer_grid-based_2019}
S.~Steyer, C.~Lenk, D.~Kellner, G.~Tanzmeister, and D.~Wollherr, ``Grid-based
  object tracking with nonlinear dynamic state and shape estimation,''
  \emph{T-ITS}, vol.~21, no.~7, pp. 2874--2893, July 2019.

\bibitem{hoermann2018dynamic}
S.~Hoermann, M.~Bach, and K.~Dietmayer, ``Dynamic occupancy grid prediction for
  urban autonomous driving: A deep learning approach with fully automatic
  labeling,'' in \emph{ICRA}, 2018.

\bibitem{piewak_fully_2017}
F.~Piewak, T.~Rehfeld, M.~Weber, and J.~M. Zollner, ``Fully convolutional
  neural networks for dynamic object detection in grid maps,'' in \emph{IV},
  2017.

\bibitem{pointnet}
R.~Q. Charles, H.~Su, M.~Kaichun, and L.~J. Guibas, ``Pointnet: Deep learning
  on point sets for 3d classification and segmentation,'' in \emph{CVPR}, 2017.

\bibitem{ilg2017flownet}
E.~Ilg, N.~Mayer, T.~Saikia, M.~Keuper, A.~Dosovitskiy, and T.~Brox,
  ``{FlowNet} 2.0: Evolution of optical flow estimation with deep networks,''
  in \emph{CVPR}, 2017.

\bibitem{nuscenes2019}
H.~Caesar, V.~Bankiti, A.~H. Lang, S.~Vora, V.~E. Liong, Q.~Xu, A.~Krishnan,
  Y.~Pan, G.~Baldan, and O.~Beijbom, ``{nuScenes}: A multimodal dataset for
  autonomous driving,'' \emph{arXiv:1903.11027}, 2019.

\bibitem{besl1992method}
P.~J. Besl and N.~D. McKay, ``Method for registration of {3-D} shapes,'' in
  \emph{T-PAMI}, vol.~14, no.~2, Feb. 1992.

\bibitem{ranjan2017optical}
A.~Ranjan and M.~J. Black, ``Optical flow estimation using a spatial pyramid
  network,'' in \emph{CVPR}, 2017.

\bibitem{nuscene-tracking}
\BIBentryALTinterwordspacing
``{nuScenes} tracking task benchmark.'' [Online]. Available:
  \url{https://www.nuscenes.org/tracking/}
\BIBentrySTDinterwordspacing

\bibitem{dellaert_factor_2017}
F.~Dellaert, M.~Kaess, \emph{et~al.}, ``Factor graphs for robot perception,''
  \emph{Foundations and Trends in Robotics}, vol.~6, no. 1-2, pp. 1--139, 2017.

\bibitem{ferrari_occlusions_2018}
E.~Ilg, T.~Saikia, M.~Keuper, and T.~Brox, ``Occlusions, motion and depth
  boundaries with a generic network for disparity, optical flow or scene flow
  estimation,'' in \emph{ECCV}, 2018.

\end{thebibliography}

\end{document}